\documentclass[10pt, a4paper]{article}
\usepackage{graphicx}
\usepackage{amsmath,amssymb} 
\usepackage{color}
\usepackage{multirow}

\usepackage{booktabs}

\begin{document}
\pagestyle{plain}

\title{Unbiased Scene Graph Generation using Predicate Similarities} 
\author{Misaki Ohashi\footnote{The University of Tokyo} \and Yusuke Matsui\footnotemark[1]}
\date{}
\maketitle

\begin{abstract}
Scene Graphs are widely applied in computer vision as a graphical representation of relationships between objects shown in images. However, these applications have not yet reached a practical stage of development owing to biased training caused by long-tailed predicate distributions.
In recent years, many studies have tackled this problem. In contrast, relatively few works have considered \textit{predicate similarities} as a unique dataset feature which also leads to the biased prediction. Due to the feature, infrequent predicates (e.g., ``\textit{parked on}", ``\textit{covered in}") are easily misclassified as closely-related frequent predicates (e.g., ``\textit{on}", ``\textit{in}"). Utilizing predicate similarities, we propose a new classification scheme that branches the process to several fine-grained classifiers for similar predicate groups. The classifiers aim to capture the differences among similar predicates in detail. We also introduce the idea of transfer learning to enhance the features for the predicates which lack sufficient training samples to learn the descriptive representations. The results of extensive experiments on the Visual Genome dataset show that the combination of our method and an existing debiasing approach greatly improves performance on tail predicates in challenging SGCls/SGDet tasks. Nonetheless, the overall performance of the proposed approach does not reach that of the current state of the art, so further analysis remains necessary as future work.
\end{abstract}

\section{Introduction}

\begin{figure}
    \centering
    \includegraphics[scale=0.23]{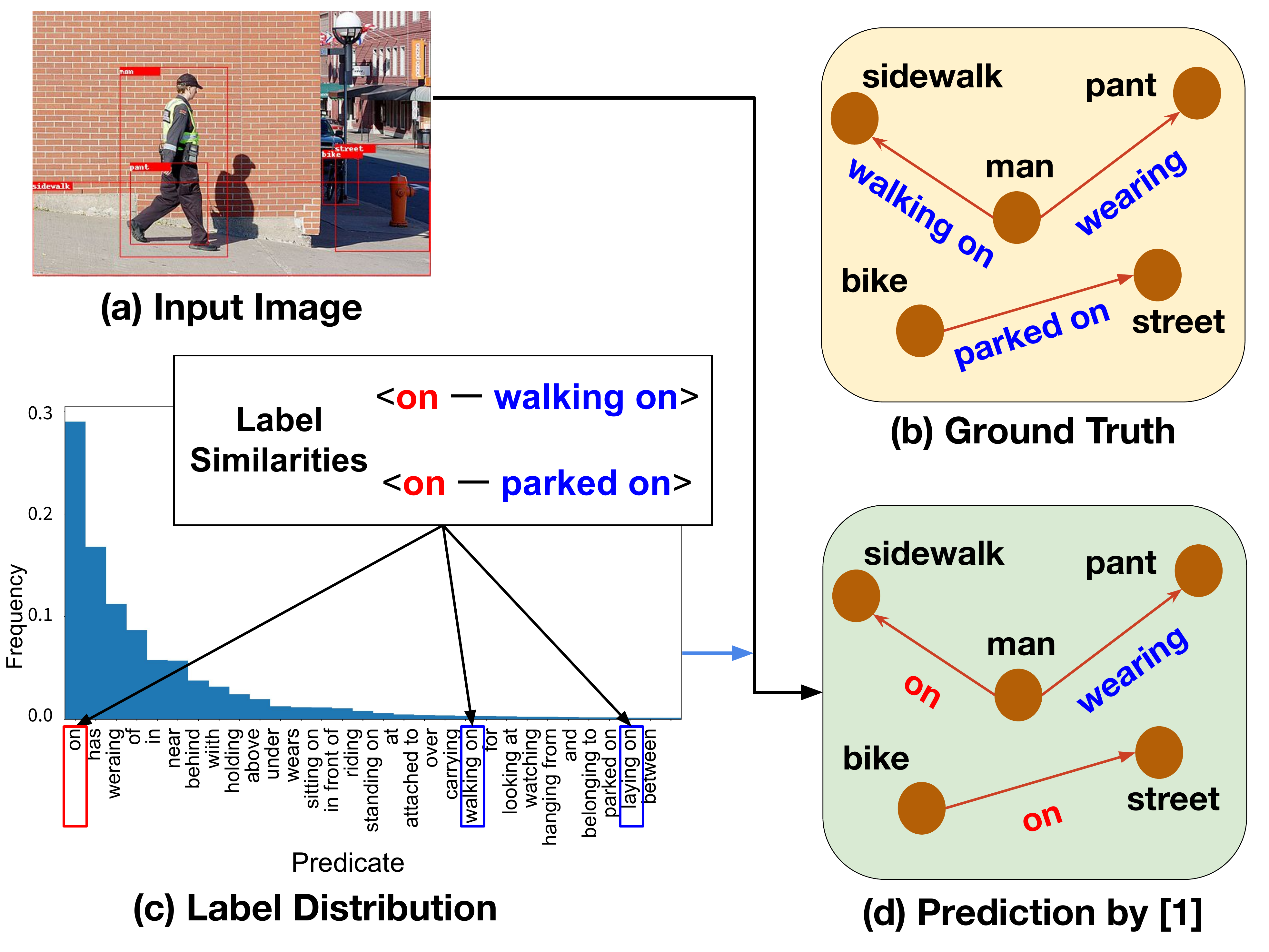}
    \caption{The class imbalance problem in scene graph generation. (a) An input image. (b) Ground-truth scene graph. (c) Frequency distribution of training samples for top-30 most frequent labels. (d) Affected by label similarities and imbalanced data distribution, the prediction results of an earlier method~\cite{motifs} misclassify some descriptive predicates as ``\textit{on}".}
    \label{fig:failure_example}
\end{figure}
Scene graphs describe objects that appear image data and their relationships in the image. Generally, scene graph generation~(SGG) is divided into three stages, including object detection, object classification, and relationship classification. Scene graphs comprehensively capture the content of image scenes. Hence, they can be applied to high-level and wide-ranging practical tasks, including visual question answering~\cite{graph_vqa,gqa2019,vqa_sgg}, image captioning~\cite{unpair_captioning,captioning_sgg,auto_captioning}, and image retrieval~\cite{image_retrieval,cross_model_image_retrieval}. 

The relationship classification stage in SGG typically involves class imbalance problems in the most widely-used Visual Genome dataset~\cite{visual_genome}. As shown in Fig.~\ref{fig:failure_example}, the number of training samples for ``\textit{on}" is about 50 times higher than ``\textit{standing on}". A model trained with such an imbalanced dataset is more likely to predict a few frequent predicates (e.g., ``\textit{on}", ``\textit{in}") against many infrequent predicates (e.g., ``\textit{lying on}", ``\textit{covered in}"). Hereafter, we refer to frequent and infrequent predicates as head and tail predicates, respectively.  

Existing unbiased methods~\cite{fewshot,exknowledge,unbiased,kt,pcpl,cogtree,ebm,bgnn,recover} have focused on the long-tailed distribution in the dataset. However, few works have focused on another unique dataset feature, \textit{predicate similarities}, which are also an important cause of the biased predictions. In contrast to general classification tasks, the dataset includes many semantically similar predicates. These similarities make distinguishing between heads and tails challenging and encourage misclassification of tail predicates as more predictable head predicates. Because head predicates are less descriptive than tail predicates, the graphs with heads are less informative and more impractical. For example, Fig.~\ref{fig:failure_example}~(b)(c) show that the behavior ``\textit{walking on}" and the state ``\textit{parked on}" are all predicted as ``\textit{on}", resulting in the ambiguous description of the image content. Scene graphs that represent limited visual information typically perform poorly in applications to high-level tasks. Therefore, SGG models should be developed to predict as specific a predicate as possible based on the subjects represented in image.

In this study, we propose a new relation predictor that utilizes the predicate similarities of the dataset. Conventional all-class classifiers consider only significant differences between dissimilar predicates. In contrast, our proposed predictor consists of several independent fine-grained classifiers, each focusing on slight differences between semantically similar predicates. The proposed approach is designed to recognize tail predicates that conventional classifiers tend to misclassify as similar head predicates.

Furthermore, inspired by earlier work~\cite{kt}, we adopt a knowledge transfer module for better representation learning. It enhances poorly learned features of tail predicates by transferring the features of heads learned with sufficient samples. In contrast to the previous method~\cite{kt}, we transfer the knowledge within similar predicates rather than all predicates. Because each fine-grained classifier targets specific similar predicates, features would be noisy if the knowledge from all predicates were incorporated, including dissimilar ones.

The contributions of this study are summarized as follows.
\begin{itemize}
    \item{We propose a method to handle the long-tail distribution and semantic similarities of predicate labels by combining a similarity-based branching scheme and a knowledge transfer module.}
    \item{The proposed method effectively improves the tails' prediction. In particular, when combined with an existing debiasing inference method, it achieved the best recall on the challenging SGCls/SGDet tasks.}
    \item{Although our approach improved the accuracy of tail labels, its overall performance was lower than the current state of the art, especially for a relatively easy task (PredCls). Further analysis remains as future work.}
\end{itemize}

\section{Related Work}

\subsection{Imbalanced Classification}\label{sec:imbalanced}
In recent years, three primary methods have been applied to perform classification tasks involving long-tailed datasets.

Data re-balancing is a classical approach that adjusts the amount of data to achieve a more balanced distribution. This method includes over-sampling for minority classes~\cite{smote,border_smote} and under-sampling for major classes~\cite{upsampling}. Over-sampling is prone to over-fitting for the tail classes, whereas undersampling discards most data,  a considerable portion of the data, which makes it difficult to apply to highly imbalanced datasets.

Cost-sensitive re-weighting assigns different loss weights based on the number of classes or samples. Commonly used methods include weighting classes proportionally to the inverse of the class frequency~\cite{inverse_reweight1,inverse_reweight2} or the inverse square root of the frequency~\cite{inverse_reweight3,inverse_reweight4}. In recent years, Cui et al.~\cite{effective_num} proposed re-weighting by an inverse effective number of samples, and Lin et al.~\cite{focalloss} introduced sample-level re-weighting.

Transfer learning involves transferring features learned from head classes with abundant samples to tail classes that are learned insufficiently. Liu et al.~\cite{oltr} introduced dynamic meta-embedding to exchange visual knowledge between heads and tails by combining a direct image feature and associated memory representations.

\subsection{Scene Graph Generation}
In the first stage of SGG, an object detector (e.g., Faster R-CNN~\cite{fasterrcnn}) detects several objects in an image. As the next step, object classification is performed after encoding the detections from the first stage into object contextual information. In most studies, the contexts are incorporated by message passing algorithms such as graph attention networks~\cite{grcnn}, LSTM~\cite{motifs}, and TreeLSTM~\cite{vctree}. Finally, the relationships among detected objects are predicted with a module similar to object classification. 

Many studies~\cite{fewshot,exknowledge,unbiased,kt,pcpl,cogtree,ebm,bgnn,recover} have proposed various methods to deal with the class imbalance problem since Chen et al.~\cite{kern} and Tang et al.~\cite{vctree} proposed the more balanced mean recall metrics. Tang et al.~\cite{unbiased} adopted a counterfactual approach in making inferences to remove a context co-occurrence bias. Chiou et al.~\cite{recover} recovered the unbiased probabilities from biased probabilities by label frequencies estimated dynamically in training. Also, recent works have adopted general ideas to address tackle long-tailed issues, as shown in Sec.~\ref{sec:imbalanced}. Li et al.~\cite{bgnn} proposed bi-level data resampling, including image-level oversampling and instance-level undersampling. Moreover, task-specific loss functions and weighting methods have also been proposed. Yan et al.~\cite{pcpl} introduced loss re-weighting by an inverse of a degree of predicate correlations. Yu et al.~\cite{cogtree} proposed a loss for a hierarchical cognitive structure to support coarse-to-fine classification. Suhail et al.~\cite{ebm} adopted a loss formulation using an energy-based model for structured learning of scene graphs. Furthermore, He et al.~\cite{kt} applied the approach of transfer learning to SGG tasks.

These recent works~\cite{fewshot,exknowledge,unbiased,kt,pcpl,cogtree,ebm,bgnn,recover} have improved SGG performance, but few studies have addressed predicate similarities in the dataset. Yan et al.~\cite{pcpl} mentioned the feature but focused on predicates having weak correlations with others, and thereby did not directly take advantage of the relationship between similar predicates. Yu et al.~\cite{cogtree} adopted a similar focus to that of the present work, but their method only considers parent-children relationships among predicates, whereas the proposed method does not limit to such hierarchical similarities.

\section{Proposed Approach}
\begin{figure*}
    \centering
    \includegraphics[scale=0.17]{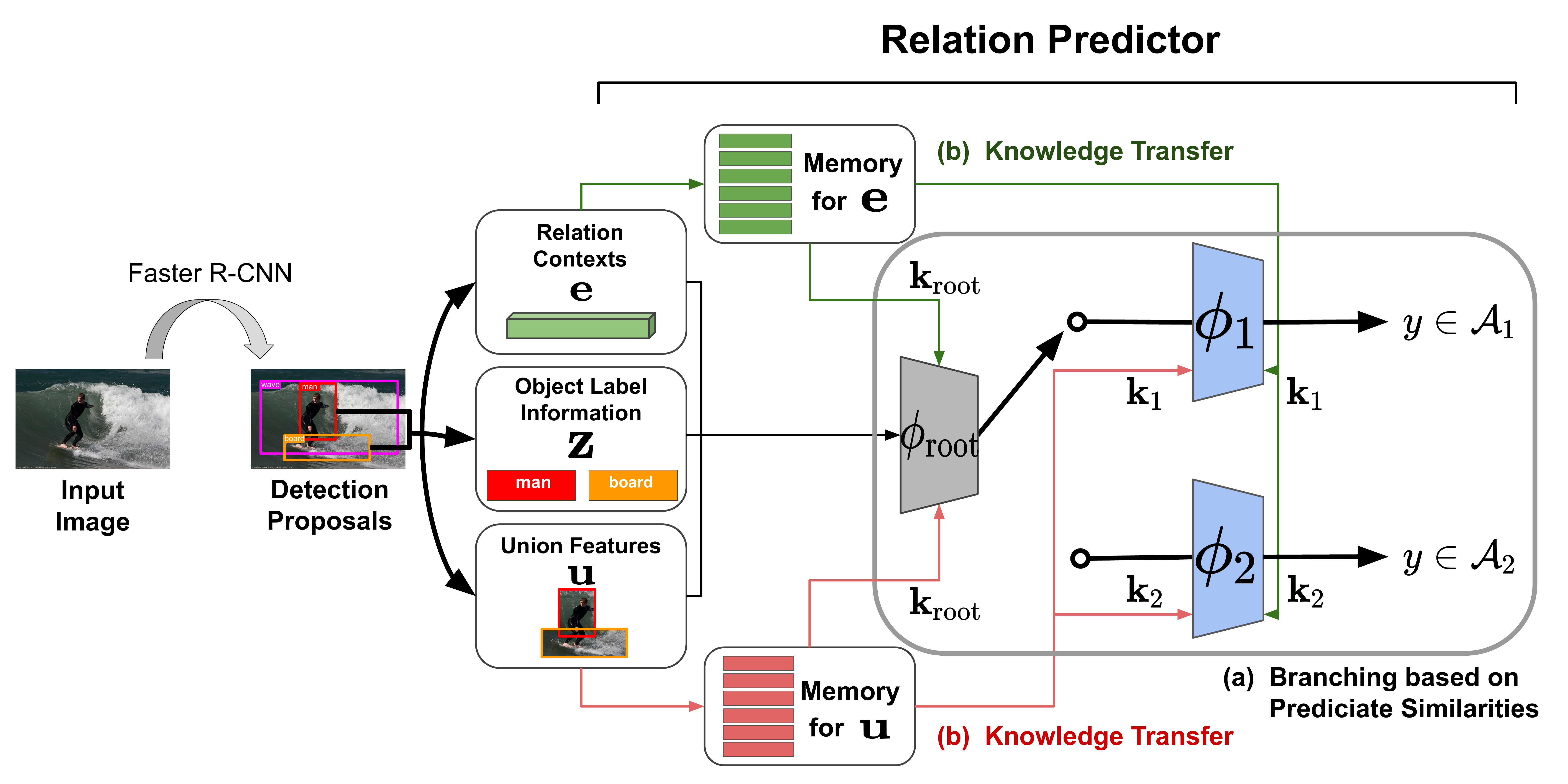}
    \caption{Model overview. First, a detector detects several objects from an input image. Using an existing SGG model, we extract the relation context $\mathbf{e}$, the embedded vector for the pairwise object labels $\mathbf{z}$, and the union box feature $\mathbf{u}$. These are all fed into the proposed relation predictor. (a) In the predictor, $\phi_\mathrm{root}$ first branches the process. If $\phi_\mathrm{root}$ chooses $\mathcal{A}_1$, $\phi_1$ then classifies predicate labels in $\mathcal{A}_1$. Otherwise, $\phi_2$ does so in $\mathcal{A}_2$. (b) $\mathbf{e}$ and $\mathbf{u}$ are enhanced before classification in each classifier. We construct knowledge features $\mathbf{k}_\mathrm{root}, \mathbf{k}_1, \mathbf{k}_2$ with memory features and then combine each feature with an original feature in each classifier.}
    \label{fig:architecture}
\end{figure*}

Scene graph generation tasks involve generating a graph representation comprising objects and the visual relationships among them shown in a given input image. In particular, we aim to address the biased relationship classification caused by imbalanced predicate distributions and semantic overlaps among the predicates. To this end, we introduce a classification strategy which focuses on predicate similarities and utilizes the idea of transfer learning. 
In this section, we first present the problem setting in Sec.~\ref{sec:prob_setting}. We then explain the details of our proposed predictor in Sec.~\ref{sec:architecture}. Fig.~\ref{fig:architecture} shows an overview of the model.

\subsection{Problem Setting}\label{sec:prob_setting}
We first detect object candidates using a standard object detector such as Faster R-CNN~\cite{fasterrcnn}. Given an image $I$, the detector outputs $N$ bounding boxes $\mathcal{B} = \{\mathbf{b}_i\}_{i=1}^N \subset \mathbb{R}^4$. Each box also includes an ROIAlign feature~\cite{roialign} and a tentative object label such as ``\textit{dog}" and ``\textit{man}". We then refine these features with a message-passing module for the final object classification and relationship classification.

Relationship classification is then performed as follows. Given a pair of bounding boxes, a relation predictor classifies the pair from a set of $A$ predicate labels (e.g., ``\textit{on}", ``\textit{in}") denoted as $\mathcal{A} = \{1,2, \dots, A\}$. Here, for each pair of bounding boxes, we have three input features, $\mathbf{e}, \mathbf{z},$ and $\mathbf{u}$ (see an example in Fig.~\ref{fig:architecture}). A $P$-dimensional pairwise relation feature $\mathbf{e} \in \mathbb{R}^P$ is obtained from the abovementioned message passing module. An embedded vector $\mathbf{z} \in \mathbb{R}^{A}$ represents the object labels of the pair. A union visual feature $\mathbf{u} \in \mathbb{R}^P$ is extracted from the union of the two boxes $\mathbf{b}_i \cup \mathbf{b}_j$. The relation predictor takes these three features as inputs and predicts a label $y \in \mathcal{A}$. Existing methods~\cite{motifs,fewshot,exknowledge,unbiased,kt,pcpl,cogtree,ebm,bgnn,recover,grcnn,vctree,kern,imp,gpsnet} use an all-class classifier here, but they may struggle to distinguish non-head predicates from semantically correlated head predicates.

In this work, we propose a plug-in relation predictor, as illustrated in Fig.~\ref{fig:architecture}. It consists of two independent fine-grained classifiers specialized for each similar predicate group. Moreover, each classifier adopts a knowledge transfer module to gain better representations for classification. In the next section, we explain the structure of the predictor further. 

\subsection{Relation Predictor Architecture}\label{sec:architecture}
Our proposed predictor consists of two components, including 1) \textit{classification by branching based on predicate similarities} (\textit{i.e.}, BRANCH) to predict relationships with finer discrimination granularity, and 2) \textit{knowledge transfer} (\textit{i.e.}, KT) to enhance tails’ representation before relationship classification. An example is shown in Fig.~\ref{fig:architecture}. We first extract $\mathbf{e}, \mathbf{z}$, and $\mathbf{u}$ from ``\textit{man}" and ``\textit{board}" objects. Given these features, our task is to predict a predicate label $y \in \mathcal{A}$.

\subsubsection{Classification by Branching based on Predicate Similarities.}\label{sec:cls_system}
First, we cluster the predicate labels into some groups based on the predicate similarities. As the simplest setting, we divide the labels~($\mathcal{A}$) into two groups~($\mathcal{A}_1 \subset \mathcal{A}$ and $\mathcal{A}_2 = \mathcal{A} \setminus \mathcal{A}_1$). According to the clustering result, we construct a classification system that first decides which group to use and then identifies a specific predicate within each group.

Before training, we measure the predicate similarities for the clustering. Considering that conventional predictor tend to confuse closely related predicates, we assume that similar predicates may be expected to exhibit similar distributions of predicted probability. Therefore we use the distance between the probability vectors as a quantitative measurement for the predicate correlations. 

For all training samples, we calculate $A$-class probability vectors with a pre-trained baseline predictor as follows.
\begin{equation}\label{eq:prob_calc}
  \left(\mathbf{e}, \mathbf{u}, \mathbf{z}\right) \mapsto \mathrm{softmax}\left(\mathbf{W}_e\mathbf{e} + \mathbf{W}_u\mathbf{u} + \mathbf{z}\right) \in \mathbb{R}^A.
\end{equation}
where $\mathbf{W}_e \in \mathbb{R}^{A \times P}$ and $\mathbf{W}_u \in \mathbb{R}^{A \times P}$ project $\mathbf{e}$ and $\mathbf{u}$ into $\mathbb{R}^A$, respectively. The $\mathrm{softmax}$ function is used to normalize the fused inputs.
We then compute an average vector for each class to obtain a representative vector. Based on the average vectors, the predicates are clustered into two groups $\mathcal{A}_1$ and $\mathcal{A}_2$ by hierarchical clustering.

During inference, we use a module which is composed from three classifiers $\phi_\mathrm{root},~\phi_1$, and $\phi_2$, as shown in Fig.~\ref{fig:architecture}~(a).  First, $\phi_\mathrm{root}$ outputs $\mathbf{p}_\mathrm{root} \in \mathbb{R}^{2}$ to branch the process. If $\mathbf{p}_\mathrm{root}[0]$ is greater than $\mathbf{p}_\mathrm{root}[1]$, we select the $\phi_1$ for the next step, suggesting that the final predicate would belong to $\mathcal{A}_1$. Otherwise, we select $\phi_2$ and the final predicate supposed to be in $\mathcal{A}_2$. Here, the fine-grained $\phi_1,~\phi_2$ output $\mathbf{p}_1 \in \mathbb{R}^{|\mathcal{A}_1|}, \mathbf{p}_2 \in \mathbb{R}^{|\mathcal{A}_2|}$ respectively. Inputs for all classifiers are $\mathbf{e}$, $\mathbf{u}$, and $\mathbf{z}$. Probability distributions $\mathbf{p}_\mathrm{root}, \mathbf{p}_1$ and $\mathbf{p}_2$ are generated in the same way with different weights as Eq.~\ref{eq:prob_calc}. 

\subsubsection{Knowledge Transfer.}\label{sec:trans_learning}
\begin{figure}
    \centering
    \includegraphics[scale=0.37]{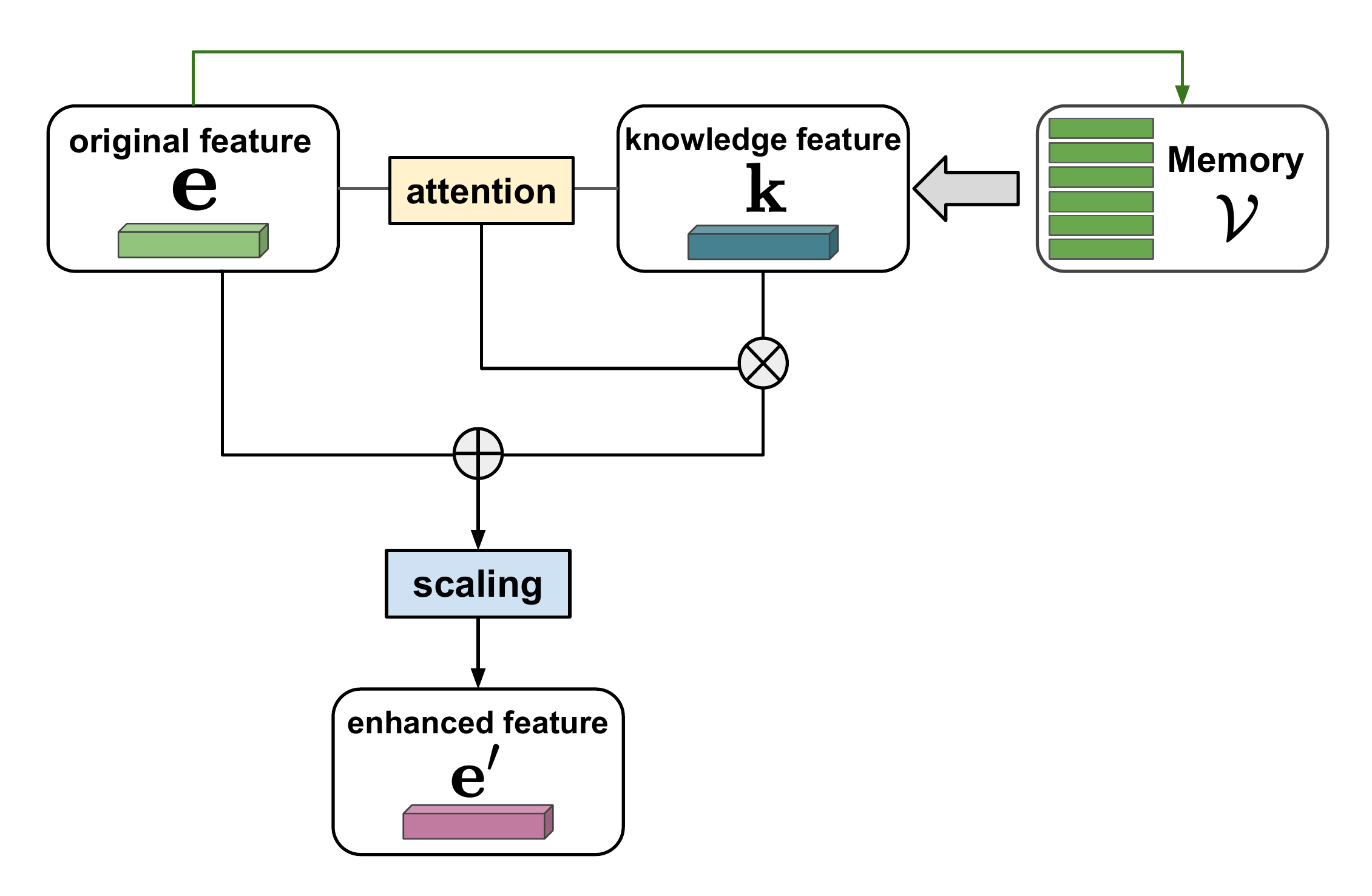}
    \caption{Knowledge Transfer Module. The original feature $\mathbf{e}$ updates the memory features $\mathcal{V}$ during training. The knowledge feature $\mathbf{k}$ is generated from $\mathcal{V}$. We then combine $\mathbf{e}$ and $\mathbf{k}$ with their attention. After scale calibration, we finally obtain the enhanced feature $\mathbf{e}^{\prime}$.}
    \label{fig:kt}
\end{figure}
Inspired by the previous work~\cite{kt}, we adopt transfer learning in $\phi_\mathrm{root},~\phi_1$, and $\phi_2$ to strengthen the input features. Among three components in the input, we enhance the contextual feature $\mathbf{e}$ and the union feature $\mathbf{u}$ before passing them down to the classifiers. Because $\mathbf{z}$ indicates a co-occurrence of the predicate labels for the given object label pair information, it is not a type of representation to transfer among predicates. In the following, we explain how $\mathbf{e}$ is enhanced in the proposed approach. Note that $\mathbf{u}$ is also refined in exactly the same way. As shown in Fig.~\ref{fig:kt}, we obtain an enriched feature $\mathbf{e}^{\prime} \in \mathbb{R}^P$ by combining an original feature $\mathbf{e}$ with a knowledge feature $\mathbf{k} \in \mathbb{R}^P$. We calculate the knowledge feature $\mathbf{k}$ from a memory $\mathcal{V}=\{\mathbf{v}_i\}_{i=1}^A$, where $\mathbf{v}_i \in \mathbb{R}^P$ is a memory feature corresponding to the $i$-th predicate label.

The original feature $\mathbf{e}$ is enhanced using the memory $\mathcal{V}$ by transferring knowledge from heads to tails. We assume that $\mathcal{V}$ represents a class centered in a feature space and captures predicate labels' generic and discriminative concepts. We follow the mechanism provided in the prior work~\cite{kt}. 

First, we compute a knowledge feature $\mathbf{k}$ from the memory $\mathbf{V} \in \mathbb{R}^{A \times P}$. Note that $\mathbf{V}$ is a matrix constructed by stacking elements in $\mathcal{V}$. Here, we obtain $\mathbf{k}$ by
\begin{equation}\label{eq:normal_knowledge}
\mathbf{k} = \mathbf{V}^\top \hat{\mathbf{p}}.
\end{equation}
where $\hat{\mathbf{p}} \in \mathbb{R}^A$ is a coefficient calculated from $\mathbf{e}$ with a linear layer followed by a softmax layer. We can consider $\hat{\mathbf{p}}$ as the preliminary prediction before the knowledge transfer.

Eq.~\ref{eq:normal_knowledge} is a form suggested by He et al.~\cite{kt} and we use it to calculate $\mathbf{k}_\mathrm{root}$ in $\phi_\mathrm{root}$. As for $\phi_1$ and $\phi_2$, we extend this and construct the knowledge with only similar predicates' memory to generate suitable representation for each fine-grained classifier. In $\phi_1$, we consider labels only in $\mathcal{A}_1$ (out of $\mathcal{A}$). This implies that we can obtain the specific knowledge for $\mathcal{A}_1$ by considering only memory features specified by $\mathcal{A}_1$.
\begin{equation}\label{eq:phi1_knowledge}
\mathbf{k}_1 = \mathbf{V}_1^\top \hat{\mathbf{p}}_1.
\end{equation}
Here, $\mathbf{V}_1 \in \mathbb{R}^{|\mathcal{A}_1| \times P}$ is a stack of $|\mathcal{A}_1|$ rows of $\mathbf{V}$ indexed by $\mathcal{A}_1$. The coefficient $\hat{\mathbf{p}}_1 \in \mathbb{R}^{|\mathcal{A}_1|}$ is calculated from $\mathbf{e}$ similarly to $\hat{\mathbf{p}}$. We can consider the final result $\mathbf{k}_1 \in \mathbb{R}^P$ to be a richer representation than a general result $\mathbf{k}$. In the same manner, $\mathbf{k}_2$ is computed with $\mathbf{V}_2 \in \mathbb{R}^{|\mathcal{A}_2| \times P}$ for $\phi_2$. Hereafter, for simplicity, $\mathbf{k}_1, \mathbf{k}_2$, and $\mathbf{k}_\mathrm{root}$ are all denoted as $\mathbf{k}$ depending on the context.

Secondly, we calculate an attention $\bf{a}$ between $\mathbf{e}$ and $\mathbf{k}$~(Eq.~\ref{eq:attention}) and combine two features with $\bf{a}$~(Eq.~\ref{eq:combine_knowledge}). Because the knowledge of tails includes a considerable amount of head information, the combined features are likely to be close to the heads' representations. This leads to difficulty distinguishing between the two. To avoid this problem, we finally set different scales for heads and tails by a maximum value $m$ in the coefficient for $V$ (e.g., $\hat{\mathbf{p}}_1$ in $\phi_1$). Given that the coefficient is considered the tentative classification vector, $m$ for heads will be larger than for tails. Let $\alpha$ be a constant value. Then, a final feature $\mathbf{e}^{\prime}$ is obtained as follows. 
\begin{eqnarray}
    &\mathrm{attention} : \left(\mathbf{x},\mathbf{y}\right) \mapsto \begin{bmatrix}
        \mathrm{max}\left(\mathrm{tanh}\left(x_1 + y_1\right), 0\right) \\
        \mathrm{max}\left(\mathrm{tanh}\left(x_2 + y_2\right), 0\right) \\ 
        \vdots
    \end{bmatrix}. \label{eq:attention_func}\\
    &\mathbf{a} = \mathrm{attention}\left(\mathbf{e}, \mathbf{k}\right).\label{eq:attention}\\
    &\mathbf{e}^{\prime} = \alpha \cdot m\left(\mathbf{e} + \mathbf{a} \odot \mathbf{k}\right). \label{eq:combine_knowledge}
\end{eqnarray}

During training, we dynamically update the memory features $\mathcal{V}$. The original feature $\mathbf{e}$ should be distant from the memory features of other predicates, whereas they should be close to the feature of the same predicate. To this end, we follow Liu et al.~\cite{oltr} and update $\mathcal{V}$ with the loss defined as
\begin{eqnarray}\label{eq:loss_memory}
    \mathcal{L}_\mathrm{mem} = \Vert \mathbf{e}-\mathbf{v}_g \Vert^2 + \gamma\cdot\mathrm{max}\left(M-\frac{1}{A}\sum_{i=1, i \neq g}^A \Vert \mathbf{e}-\mathbf{v}_i \Vert,  0\right).
\end{eqnarray}
Let $g \in \mathcal{A}$ denote a ground-truth predicate label for $\mathbf{e}$. $\gamma$ is a constant scalar to balance the terms, and $M$ is a constant margin.  Intuitively, the first term pulls $\mathbf{e}$ and the memory feature of $g$ together, and the second separates $\mathbf{e}$ from the memory features of the other predicates. In training, we initialize $\mathcal{V}$ with an average $\mathbf{e}$ for each class which is computed by a pre-trained baseline model with all training samples. Then, we dynamically improve it with $\mathcal{L}_\mathrm{mem}$.

\subsubsection{Loss Function.}
To train our relation predictor, we calculate a classification loss $\mathcal{L}_\mathrm{rel}$. As in Eq.~\ref{eq:loss_memory}, we define $g$ as a correct predicate label. In $\phi_1$, the output $\mathbf{p}_1$ from our model is calculated as follows.
\begin{equation}
    \mathbf{p}_1 = \mathrm{softmax}\left(\mathbf{W}^1_e \mathbf{e}^{\prime} + \mathbf{W}^1_u \mathbf{u}^{\prime}+\mathbf{W}^1_z \mathbf{z}\right).
\end{equation}
$\mathbf{W}^1_e \in \mathbb{R}^{|\mathcal{A}_1| \times P}$, $\mathbf{W}^1_u \in \mathbb{R}^{|\mathcal{A}_1| \times P}$ and $\mathbf{W}^1_z \in \mathbb{R}^{|\mathcal{A}_1| \times A}$ map the inputs to $\mathbb{R}^{|\mathcal{A}_1|}$. In the same way as $\mathbf{e}^{\prime}$, $\mathbf{u}^{\prime}$ denotes a new union feature after knowledge transfer. The classification loss in $\phi_1$ is defined as $\mathcal{L}^1_\mathrm{rel}$. It includes cross-entropy loss for the final output $\mathbf{p}_1$, as well as auxiliary cross-entropy losses for individual predictions from three inputs: $\mathbf{W}^1_e \mathbf{e}^{\prime}$, $\mathbf{W}^1_u \mathbf{u}^{\prime}$, $\mathbf{W}^1_z\mathbf{z}$. Furthermore, we add cross-entropy loss for the coefficient $\hat{\mathbf{p}}_1$ in Eq.~\ref{eq:phi1_knowledge} to learn an effective way to combine the memory features.
Therefore, $\mathcal{L}^1_\mathrm{rel}$ is calculated as follows.
\begin{eqnarray}
    \mathcal{L}^1_\mathrm{rel} &=& \mathrm{CE}\left(\mathbf{p}_1, g\right) + \mathrm{CE}\left(\hat{\mathbf{p}}_1, g\right)  \nonumber \\ 
    &+& \mathrm{CE}\left(\mathbf{W}^1_e \mathbf{e}^{\prime}, g\right)
    + \mathrm{CE}\left(\mathbf{W}^1_u \mathbf{u}^{\prime}, g\right) + \mathrm{CE}\left(\mathbf{W}^1_z\mathbf{z}, g\right)
\end{eqnarray}
where $\mathrm{CE}$ denotes cross-entropy. Losses in $\phi_\mathrm{root}$ and $\phi_2$ are defined as $\mathcal{L}_\mathrm{rel}^\mathrm{root}$ and $\mathcal{L}_\mathrm{rel}^2$, which are computed in the same way. We sum all the classification losses to gain $\mathcal{L}_\mathrm{rel}$.
\begin{equation}\label{eq:loss_rel}
    \mathcal{L}_\mathrm{rel} = \mathcal{L}^1_\mathrm{rel} + \mathcal{L}^2_\mathrm{rel} + \mathcal{L}^\mathrm{root}_\mathrm{rel}
\end{equation}
The overall loss function is as follows.
\begin{equation}\label{eq:loss_all}
    \mathcal{L} = \mathcal{L}_\mathrm{obj} + \mathcal{L}_\mathrm{rel} + \lambda\mathcal{L}_\mathrm{mem}
\end{equation}
where $\mathcal{L}_\mathrm{obj}$ is a object detection loss in Faster R-CNN~\cite{fasterrcnn} and $\mathcal{L}_\mathrm{mem}$, $\mathcal{L}_\mathrm{rel}$ are computed as Eq.~\ref{eq:loss_memory}, Eq.~\ref{eq:loss_rel}. $\lambda$ is a parameter to balance the contribution.

\section{Experiments}
\subsection{Dataset Details}
We evaluate our proposed method on the Visual Genome dataset~\cite{visual_genome}. Because the original dataset has noisy annotations, we follow
the same split protocol as Xu et al.~\cite{imp} to use the 150 most frequent object categories and 50 predicates. The data is divided into a training set and a testing set. The training set includes 70\% of the images, including $5,000$ images for validation, and the testing set comprises the remaining 30\% of the images.  

\subsection{Evaluation Protocol}
We follow three standard setups for evaluation.
\begin{itemize}
    \item{Predicate Classification (PredCls) predicts the predicate labels for a set of object pairs given ground-truth bounding boxes and object labels.}
    \item{Scene Graph Classification (SGCls) predicts the object labels for ground-truth bounding boxes and the predicate labels for object pairs.}
    \item{Scene Graph Generation (SGDet) predicts bounding boxes, object labels, and predicate labels for object pairs, taking only an image as input.}
\end{itemize}
To evaluate the performance of each relationship without the results of being dominated by the performance of head predicates, we follow Tang et al.~\cite{unbiased} to adopt mean recall@K (mR@K) as an evaluation metric. For a more detailed analysis, we also calculate mean recalls for three predicate groups, including top (10 most frequent labels), middle (mid-25 labels), and bottom (15 least frequent labels). Because the ten most frequent predicates account for a relatively high percentage of training samples, we consider middle and bottom-15 as essential metrics for tail predicates.

\subsection{Implementation Details}
Following Tang et al.~\cite{unbiased}, we use a pre-trained Faster R-CNN~\cite{fasterrcnn} with a ResNext-101-FPN~\cite{resnext} backbone as an object detector and freeze the weights in scene graph generation training. The batch size is set to be 12 for PredCls and 4 for SGCls and SGDet. We use the SGD optimizer with an initial learning rate of 0.01 after 500 warm-up iterations. We set $\alpha$ = 10 in Eq.~\ref{eq:combine_knowledge}, $\gamma$ = 0.01 and $M$ = 80 in Eq.~\ref{eq:loss_memory}. $\lambda$ in Eq.~\ref{eq:loss_all} is initialized as 1 for PredCls, 0.1 for SGCls and SGDet.

We use Motifs model~\cite{motifs} to extract contextual representations and classifying object labels. Before training our predictor, we train the baseline SGG model (Motifs + a conventional relation predictor for all predicate labels). As explained in Sec.~\ref{sec:architecture}, we use the pretrained baseline model to calculate average probability distributions to cluster predicates and mean representations to initialize the memory $\mathcal{V}$. To train our relation predictor, we replace the baseline relation predictor with ours and fine-tune only the  predictor. 

We experiment with two cases, using only our method itself and combining our method with existing unbiased inference method. In this experiment, we use TDE~\cite{unbiased} for the second case as the simplest among existing methods.  TDE~\cite{unbiased} is a method designed to remove a side effect of scene information and language prior from biased probabilities. Our method is designed to perform unbiased training, so we aim to further improve its performance through the mutual effect of different approaches.

\begin{figure}[t]
    \centering
    \includegraphics[scale=0.6]{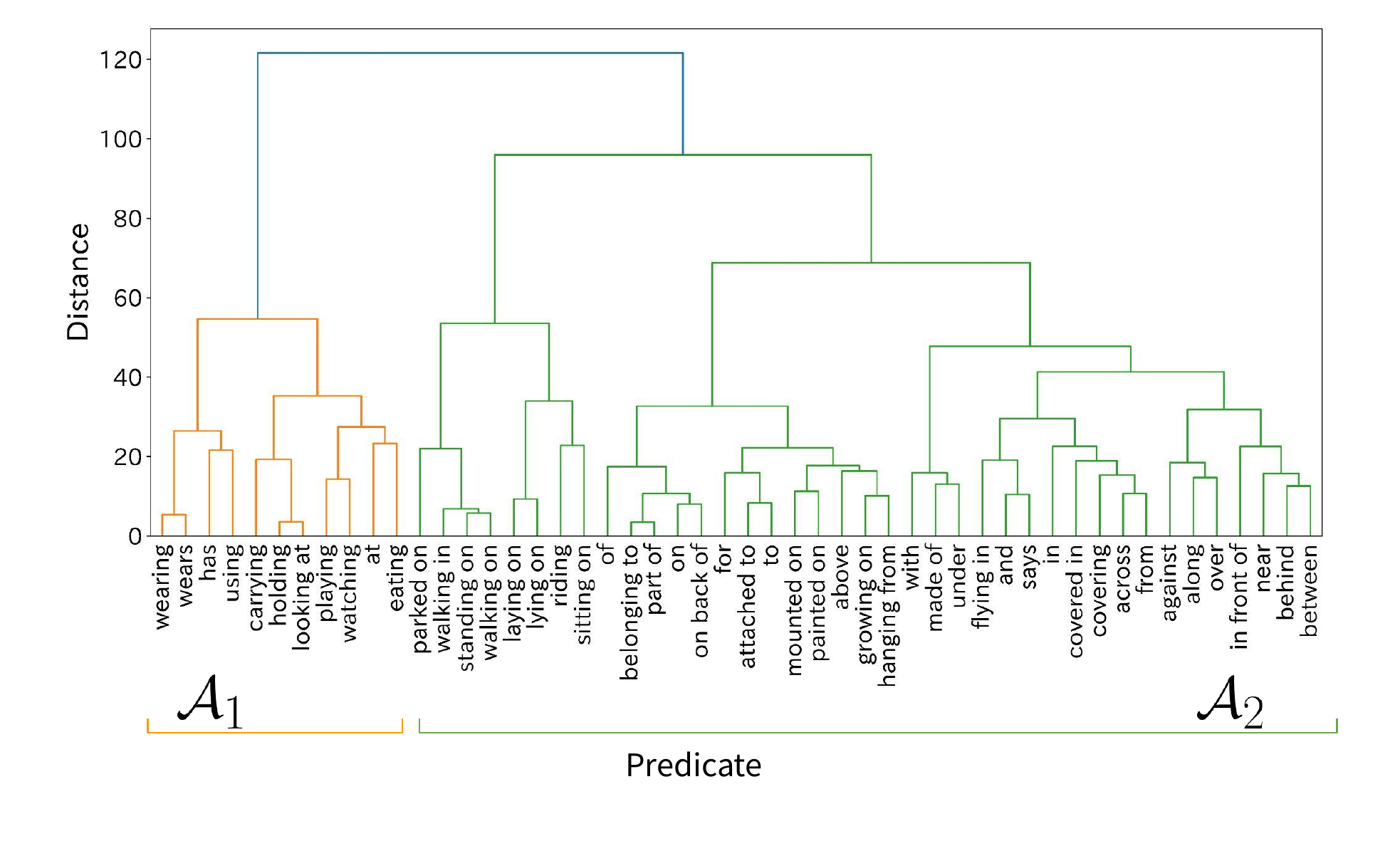}
    \caption{Clustering result in PredCls task.}
    \label{fig:pre_clustering}
\end{figure}

\subsection{Clustering Details}
Before showing the main result, we present the result of hierarchical clustering on the predicates in PredCls in Fig.~\ref{fig:pre_clustering}~(tree diagram). When we divide the predicates in two groups, one (orange) exhibits predicates similar to ``\textit{has}", and the other (green) has predicates which are semantically close to ``\textit{on}" and ``\textit{in}". They each correspond to $\mathcal{A}_1$ and $\mathcal{A}_2$. Clustering results are similar on SGCls and SGDet tasks.

\begin{table}[t]
\begin{center}
    \caption{Performance comparison in mR@K on Visual Genome dataset~\cite{visual_genome}.}
  \label{tab:comp}
  \resizebox{\textwidth}{!}{
    \begin{tabular}{@{}l r r r r r r r r r @{}} \toprule
   & \multicolumn{3}{c}{PredCls} & \multicolumn{3}{c}{SGCls}  & \multicolumn{3}{c}{SGDet}  \\
    \cmidrule{2-4} \cmidrule(l){5-7} \cmidrule(l){8-10}
       Method & mR@20 & @50 & @100 & mR@20 & @50 & @100 & mR@20 & @50 & @100 \\ \midrule
       MOTIFS~\cite{motifs,unbiased} & 11.6 & 14.6 & 15.8 & 6.1 & 7.5 & 7.9 & 4.0 & 5.4 & 6.6 \\
       ~w/ TDE~\cite{unbiased} & 17.6 & 24.4 & \underline{28.2} & 8.4 & 11.6 & 13.2 & 5.6 & 7.9 & 9.5 \\
       ~w/ PCPL~\cite{pcpl,recover} & 17.6 & 21.7 & 23.5 & 6.7 & 8.5 & 9.1 & 5.6 & 7.8 & 9.1 \\
       ~w/ CogTree~\cite{cogtree} & \bf{23.0} & \bf{28.5} & \bf{30.8} & \bf{13.0} & \bf{15.6} & \bf{16.7} & \bf{8.6} & \bf{11.5} & \bf{13.7} \\
       ~w/ EBM~\cite{ebm} & 13.3 & 16.9 & 18.4 & 7.1 & 8.8 & 9.3 & 5.2 & 7.2 & 8.5 \\
       ~w/ DLFE~\cite{recover} & \underline{20.8} & \underline{25.7} & 27.4 & 11.4 & 13.8 & 14.5 & \underline{8.0} & \underline{10.9} & \underline{12.6} \\ \midrule
       ~w/ Ours & 14.3 & 18.0 & 19.4 & 7.4 & 9.3 & 10.1 & 4.8 & 6.6 & 8.1 \\
       ~w/ Ours+TDE~\cite{unbiased} & 15.8 & 19.8 & 21.3 & \underline{12.4} & \underline{15.5} & \bf{16.7} & 7.2 & 10.1 & 12.4 \\
        \bottomrule
     \end{tabular}
     }
\end{center}
\end{table}


\begin{table}[t]
\begin{center}
    \caption{Per-group results on Visual Genome dataset~\cite{visual_genome}. }
    \label{tab:pergroup}
    \resizebox{\textwidth}{!}{
     \begin{tabular}{@{}l r r r r r r r r r  @{}} \toprule
      & \multicolumn{3}{c}{PredCls} & \multicolumn{3}{c}{SGCls}  & \multicolumn{3}{c}{SGDet}  \\
      \cmidrule{2-4} \cmidrule(l){5-7} \cmidrule(l){8-10}
       Method & top & middle & bottom & top & middle & bottom & top & middle & bottom \\ \midrule
       MOTIFS~\cite{motifs,unbiased} & \bf{56.5} & 8.8 & 0.3 & 32.3 & 2.9 & 0 & 27.8 & 2.1 & 0 \\
       ~w/ TDE~\cite{unbiased} & 48.0 & \bf{36.2} & 1.8 & 27.1 & 15.3 & 0.3 & 20.3 & 10.5 & 0.5 \\
       ~w/ PCPL~\cite{pcpl,recover} & 54.3 & 25.0 & 0.4 & 32.2 & 5.4 & 0 & 28.9 & 6.6 & 0.2 \\
       ~w/ CogTree~\cite{cogtree} & 47.1 & 35.0 & 13.0 & 28.1 & \bf{18.6} & 5.8 & 24.6 & \bf{15.3} & 3.8 \\
       ~w/ EBM~\cite{ebm} & 55.8 & 13.6 & 1.4 & \bf{32.9} & 5.0 & 0.8 & \bf{30.1} & 4.6 & 0.6 \\
       ~w/ DLFE~\cite{recover} & 47.1 & 26.6 & \bf{15.5} & 27.6 & 15.1 & 4.8 & 24.0 & 12.5 & 5.4 \\
       \midrule
       ~w/ Ours &  54.1 & 15.7 & 2.4 & 32.2 & 6.6 & 1.1 & 27.8 & 4.5 & 1.0 \\
       ~w/ Ours+TDE~\cite{unbiased} & 36.1 & 25.0 & 5.2 & 24.0 & 17.6 & \bf{10.5} & 23.0 & 12.0 & \bf{6.0} \\
        \bottomrule
     \end{tabular}}
\end{center}
\end{table}


\begin{table}[t]
\begin{center}
  \caption{Ablation study of model structure in terms of  mR@100 on Visual Genome dataset. Models are all trained with our codebase using Motifs~\cite{motifs} as an SGG backbone. BRANCH means branching based on predicate similarities and KT denotes knowledge transfer module. The baseline model is shown on the top.}
  \begin{tabular}{@{} l c c  c  c  c @{}} \toprule
   & \multicolumn{2}{c}{Method} & \multicolumn{3}{c}{Evaluation criteria} \\
  \cmidrule(r){2-3} \cmidrule{4-6}
  & BRANCH & KT & PredCls & SGCls & SGDet \\ 
    \midrule
    \multirow{4}{*}{w/o TDE} & $\times$ & $\times$ & 15.8 & 7.9 & 6.6 \\
    & $\checkmark$ & $\times$ & 17.3 & 10.0 & 8.3 \\
    & $\times$ & $\checkmark$ & 17.1 & 9.0 & 7.2  \\
    & $\checkmark$ & $\checkmark$ & 19.4 & 10.1 & 8.1 \\ \midrule
    \multirow{4}{*}{w/ TDE} & $\times$ & $\times$ & 28.2 & 13.2 & 9.5 \\
    & $\checkmark$ & $\times$ & 30.7 & 15.3 & 11.7 \\
    & $\times$ & $\checkmark$ & \bf{30.9} & 15.7 & 10.8 \\
    & $\checkmark$ & $\checkmark$ & 21.3 & \bf{16.7} & \bf{12.4} \\ 
    \bottomrule
  \end{tabular}
  \label{tab:ablation}
\end{center}
\end{table}

\begin{figure}[t]
    \centering
    \includegraphics[scale=0.24]{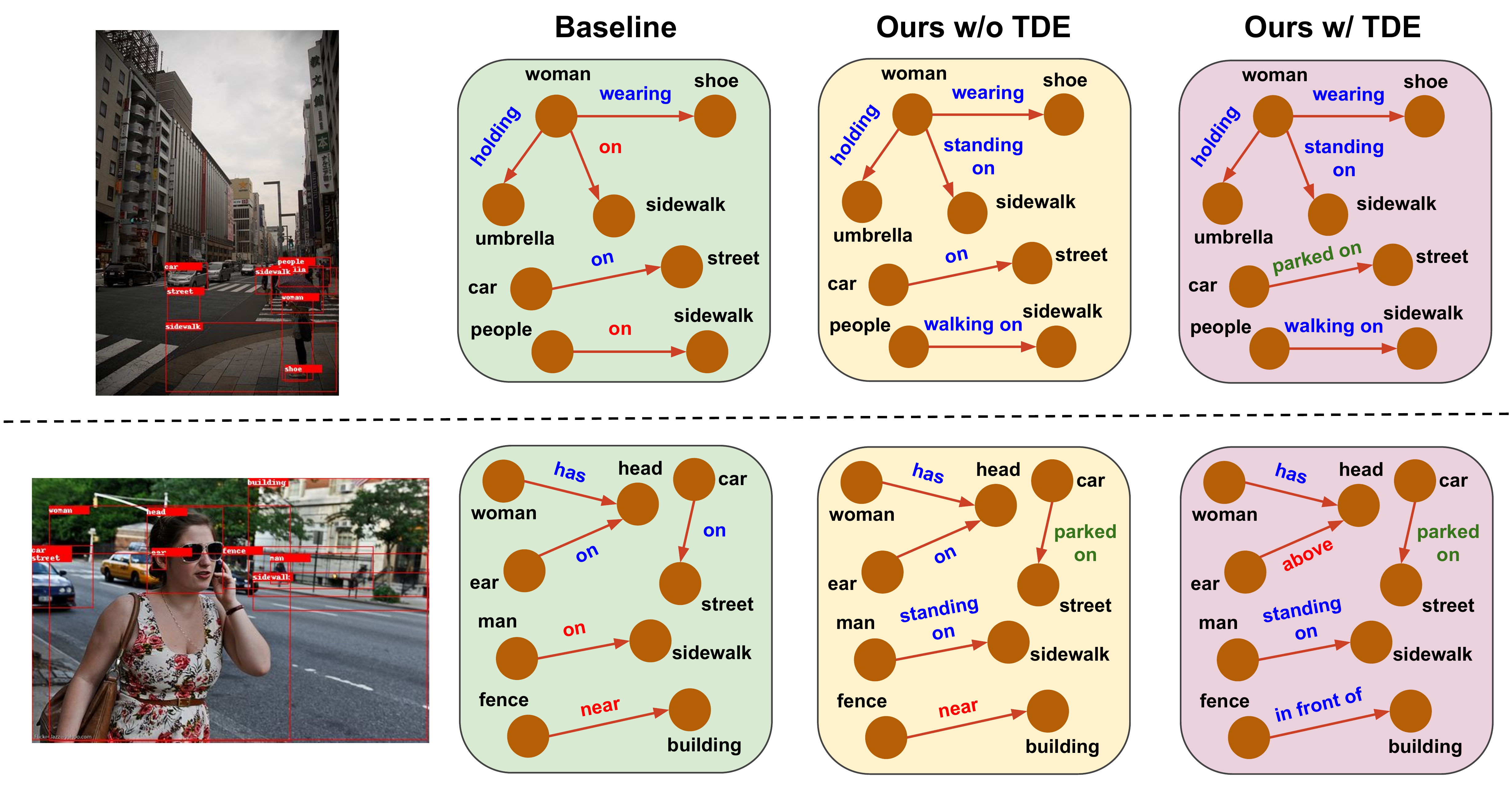}
    \caption{Scene Graph outputs in PredCls from Motifs~\cite{motifs}. Relationships in red, blue, and green mean incorrect, correct, and different from GT but reasonable predictions, respectively.}
    \label{fig:sgg_example}
\end{figure}

\subsection{Comparison with State-of-the-Art Methods}
For a fair comparison, we compared our approach with the MOTIFS~\cite{motifs}-based debiasing methods: TDE~\cite{unbiased}, PCPL~\cite{pcpl}, CogTree~\cite{cogtree}, EBM~\cite{ebm}, DLFE~\cite{recover}. We retrain these models with the same batch size as ours using the existing codebase.

Table~\ref{tab:comp} shows the overall performance (mean Recall). When we use our method by itself,the model does not outperform most of the other approaches in all the tasks. When TDE is added, though our method still perform more poorly than CogTree and DLFE in PredCls, it achieves comparable results with the current state-of-the-art (CogTree) in SGCls and SGDet. Considering these results, CogTree and DLFE may be considered the more powerful methods overall, but our method with TDE exhibits performance close to theirs. 

Table~\ref{tab:pergroup} shows per-group results. Our proposed approach outperforms most of the previous works in bottom group, which is the group of the least frequent predicates. Although bottom recalls in CogTree and DLFE are higher than in the proposed approach, they noticeably decrease top recall. When combined with TDE, our proposed approach achieves the best bottom recall in the challenging SGCls/SGDet tasks. This indicates that our focus on predicate similarities improves tails' accuracy in close settings to the real world. Nonetheless, it still performs more poorly than CogTree and DLFE in PredCls and top recall significantly drops from the baseline in all the settings because of TDE.

In PredCls, unlike SGCls and SGDet, our method with TDE shows lower mean recall than some existing works both as a whole and as a group. In this task, recall improvement by TDE is relatively more minor than in the other two tasks. Concretely, TDE increases mR@100 by 4-6~\% in the SGCls and SGDet, but only by 1~\% in PredCls. Given that, we assume that there might be conflicts between our method and TDE in the PredCls. TDE supposes that the initial classification is biased by the imbalanced label distribution and corrects it with a counterfactual approach. Our method improves tail predicates’ recall (middle and bottom) by an average of about 5\% over the baseline, implying that it already mitigates the population-based predictions to some extent. Moreover, TDE could exclude helpful bias from the union features and the pairwise object label information. Since PredCls provides ground-truth bounding boxes and object labels, they are relatively informative compared to the other two tasks. Thus, they may give ``good bias” to narrow down the candidates of the predicate labels.

\subsection{Ablation Study}\label{sec:ablatiion}
We evaluate the importance of two components in our model: \textit{classification by branching based on predicate similarities}~(BRANCH), and \textit{knowledge transfer}~(KT). As shown in Table \ref{tab:ablation}, we incrementally add each component to check their effectiveness. In the case without TDE, the recall increases sequentially as we add the components. The model with both BRANCH and KT improves on the baseline by 3.6\%, 2.2\%, 1.5\% in PredCls, SGCls, and SGDet, respectively. These results demonstrate that each proposed component functions effectively in our model. 

When applying TDE during inference, we achieve the best recall with BRANCH and KT jointly added in SGCls and SGDet. There are relative improvements in mR@100 of 8.8\% and 5.8\% from the baselines, respectively. In PredCls, on the other hand, the model with each component alone shows better performance than with all the components. The model with BRANCH alone and that with KT alone outperform ours by 9.4\% and 9.6\%, respectively. As mentioned, TDE exhibits a smaller effect when the initial classification is balanced. Without TDE,  combining two components improves the recall 2\% better than applying each independently. Therefore, we consider that TDE could not maximize its effectiveness for our entire model.

\subsection{Qualitative Results}
Fig.~\ref{fig:sgg_example} visualizes the qualitative results gained from the baseline and our models trained in PredCls. We follow Chiou et al.~\cite{recover} to indicate the reasonable but non-GT relationships like ``\textit{parked on}" in green. For both images, output scene graph becomes more appropriate and detailed in the baseline, our approach without TDE and ours with TDE, in that order. As for the first example, some relationships are lumped into ``\textit{on}" in the baseline. In contrast, our models express them as specific actions such as ``\textit{standing on}" and ``\textit{walking on}", leading to the generation of the more informative scene graphs. When adding TDE in inference, the relationship between ``\textit{car}" and ``\textit{street}" is replaced with ``\textit{on}" to ``\textit{parked on}", and the graph further reflects the image content. In the second example, the middle graph improves the left one by clearly describing the man’s action as ``\textit{standing on}" and the car’s state as ``\textit{parked on}". The right image, the vague location relationship ``\textit{near}" is then replaced with ``\textit{in front of}".

\section{Limitation and Future Works}
Our method with TDE outperforms the existing works in SGCls and SGDet tasks, but not in PredCls. Furthermore, TDE causes significant reduction of top recall. Therefore, our proposed method and TDE might not be the best combination. To find the optimal combination, further experiments with the other debiasing methods such as DLFE would be required.

Furthermore, we would like to try various model architectures with different numbers of predicate groups. We split predicates into two groups as the most straightforward setup in this experiment. However, it might be possible to make each classifier more specified by dividing predicates more finely based on similarities. Taking a balance between branching accuracy in $\phi_\mathrm{root}$ and classification accuracy in the other classifiers, the classification system should be further optimized in future work.

\section{Conclusion}
In the study, We present a relationship classification method for unbiased SGG, based on the predicate similarities, which have thus far received relatively little attention so far. We first develop branches based on the similarities to learn the difference among similar predicates in detail. Moreover, we adopt transfer learning to obtain better features for tail predicates that lack training samples. The results of experiments on the Visual Genome dataset shows that our strategy with TDE improves the recall of tail samples better than other state-of-the-art methods on SGCls and SGDet tasks. This result indicates that focusing on the label similarities and inference adjustment are mutually effective for better tail prediction. Though our method improves the accuracy of predicting tail predicates, further improvement in terms of overall performance remains necessary.

\bibliographystyle{splncs}
\bibliography{ref}

\end{document}